\documentclass{article}

\usepackage{microtype}
\usepackage{graphicx}
\usepackage{booktabs} 

\usepackage{hyperref}

\usepackage[accepted]{icml2024}

\usepackage{amsmath}
\usepackage{amssymb}
\usepackage{mathtools}
\usepackage{amsthm}

\usepackage[capitalize,noabbrev]{cleveref}

\theoremstyle{plain}

\theoremstyle{definition}

\theoremstyle{remark}

\usepackage[textsize=tiny]{todonotes}

\usepackage{times}  
\usepackage{helvet}  
\usepackage{courier}  
\usepackage{natbib}  
\usepackage{caption} 

\usepackage{newfloat}
\usepackage{listings}
\usepackage{tabularx}
\usepackage{multirow}
\usepackage{subfig}
\usepackage{adjustbox}

\usepackage[T1]{fontenc}

\usepackage{xcolor}
\usepackage{colortbl}

\usepackage{makecell} 
\usepackage{ulem}     

\usepackage{hyperref}

\newcommand{\mcal}{\mathcal}

\makeatletter
\newcommand{\thickhline}{%
	\noalign {\ifnum 0=`}\fi \hrule height 1.5pt
	\futurelet \reserved@a \@xhline
}
\makeatother
\usepackage{bm}

\icmltitlerunning{LLVD: LSTM-based Explicit Motion Modeling in Latent Space for Video Denoising}

\begin{document}

\twocolumn[
\icmltitle{LLVD: LSTM-based Explicit Motion Modeling in Latent Space for Blind Video Denoising}



\icmlsetsymbol{equal}{*}

\begin{icmlauthorlist}
\icmlauthor{Loay Rashid}{equal,yyy}
\icmlauthor{Siddharth Roheda}{equal,yyy}
\icmlauthor{Amit Unde}{equal,yyy}
\end{icmlauthorlist}

\icmlaffiliation{yyy}{Samsung Research Institute, Bangalore}

\icmlcorrespondingauthor{Loay Rashid}{loayrashid1@gmail.com}
\icmlcorrespondingauthor{Siddharth Roheda}{sid.roheda@samsung.com}

\icmlkeywords{Machine Learning, ICML}

\vskip 0.3in
]




\printAffiliationsAndNotice{\textsuperscript{*}Equal contribution }

\begin{abstract}

Video restoration plays a pivotal role in revitalizing degraded video content by rectifying imperfections caused by various degradations introduced during capturing (sensor noise, motion blur, etc.), saving/sharing (compression, resizing, etc.) and editing. This paper introduces a novel algorithm designed for scenarios where noise is introduced during video capture, aiming to enhance the visual quality of videos by reducing unwanted noise artifacts. We propose the Latent space LSTM Video Denoiser (LLVD), an end-to-end blind denoising model. LLVD uniquely combines spatial and temporal feature extraction, employing Long Short Term Memory (LSTM) within the encoded feature domain. This integration of LSTM layers is crucial for maintaining continuity and minimizing flicker in the restored video. Moreover, processing frames in the encoded feature domain significantly reduces computations, resulting in a very lightweight architecture. LLVD's blind nature makes it versatile for real, in-the-wild denoising scenarios where prior information about noise characteristics is not available. Experiments reveal that LLVD demonstrates excellent performance for both synthetic and captured noise. Specifically, LLVD surpasses the current State-Of-The-Art (SOTA) in RAW denoising by 0.3dB, while also achieving a 59\% reduction in computational complexity.


\end{abstract}

\section{Introduction}
\label{sec:intro}

\begin{figure}[t]
	\centering
	\includegraphics[width=\linewidth]{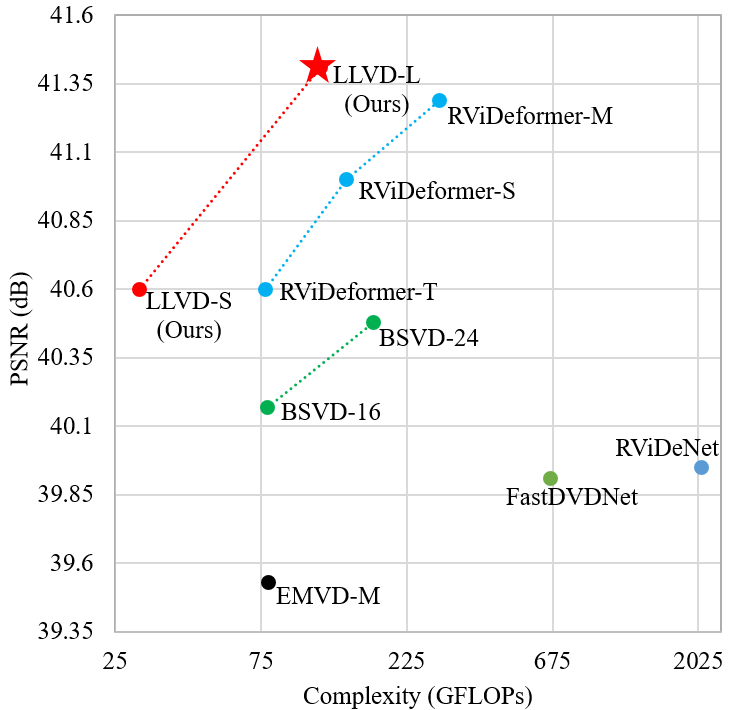}
	\caption{Comparison of PSNR (dB) and computational complexity (GFLOPs) of models on the sRGB CRVD testset. Compared to existing methods, our models (LLVD-S/L) achieve State-Of-The-Art denoising performance with significantly lower complexity.}
	\label{fig:graph}
\end{figure}

Despite significant progress in camera sensor technology, captured images and videos suffer from degradations in perceptual quality due to the presence of noise. The noise in captured images/videos is complex and stochastic in nature due to randomness present in the photon counting process and imprecision in the hardware circuitry. This noise causes severe degradation in image quality, especially in unconstrained scenarios such as low-light environments and poor camera sensors. Therefore, denoising, the process of reducing noise artifacts and restoring fine details in images/videos, is an important problem in the field of computer vision.

While the field of image denoising has been extensively explored \cite{img_den_survey2}, the research community is showing an increasing interest in video denoising. Video denoising differs from image denoising in two key aspects: (1) the temporal correlation between adjacent video frames holds the potential to significantly boost denoising performance and (2) processing the added temporal dimension poses a major computational challenge. Therefore, there is a need for video denoising algorithms which efficiently model the temporal relations present between video frames to boost denoising performance, while simultaneously requiring minimal resources to accomodate computationally constrained systems such as smartphones.

In recent years, a variety of video denoising algorithms have been proposed, ranging from motion-estimation based approaches \cite{dvd} to emerging learning based approaches \cite{fastdvd}. From an extensive examination of pre-existing methods, it becomes evident that classical methods based on either frame averaging, spatio-temporal patch based processing \cite{vbm4d}, non-local-means filtering \cite{nlm}, or optical flow \cite{flow1} demand huge processing power, making the extension of these methods to real-time applications extremely challenging. 

The advent of Convolutional Neural Networks (CNNs) has marked a significant turning point in this domain. Deep-CNN based algorithms for video denoising have shown significant improvements in performance and are a promising direction of research. CNN-based denoising approaches primarily exploit temporal redundancy either through implicit or explicit motion modeling. 

Explicit motion modeling often relies on optical flow and motion estimation and/or compensation techniques \cite{meshflow}. Despite their efficacy, explicit motion modeling techniques are very computationally intensive, restricting their practical applicability. In this context, a number of algorithms have emerged that capitalize on the strong correlation between neighbouring frames by utilizing multi-stage cascade architectures \cite{udvd, rvidenet, fastdvd}. While this strategy reduces complexity to some extent, it suffers from noticeable reductions in denoising performance. Additionally, these multi-stage methods are still not lightweight enough for denoising on consumer devices.


The inferior performance of implicit motion modeling methods may be attributed to their limited capability in exploiting the temporal correlations between frames. This challenge can be addressed by employing Long Short-Term Memory (LSTM) networks. Known for their excellence in handling sequential data, LSTM networks are ideally suited for modeling the spatio-temporal relationships in video frames, offering a promising solution to enhance video denoising performance.


\textbf{Contributions:}
In this paper, we present a novel Latent Space LSTM Video Denoiser (LLVD) dedicated to the vital task of blind video denoising. Our primary contributions include:
\begin{itemize}
    \item An innovative architecture that seamlessly integrates LSTM layers in the encoded latent space of individual video frames. This approach facilitates the acquisition of temporal relationships between consecutive frames, while also capturing elusive long-term dependencies that play a crucial role in the restoration process. 
    \item A light-weight blind denoising approach, which makes video denoising on consumer smartphones (and other embedded devices) a feasible reality. The inclusion of latent LSTMs leads to an astounding reduction in the computations needed for denoising a single frame when compared against SOTA methods.
    \item We provide a detailed ablation study to highlight and justify the impact of LSTMs in the latent space and demonstrate that they serve as a better alternative to independently using ConvLSTMs or spatial Encoders.
    \item LLVD demonstrates excellent denoising performance on both synthetic Gaussian noise and real-world captured noise benchmark datasets, achieving SOTA performance without prior information of the noise characteristics, while requiring significantly lower computations.

    

\end{itemize}

%

\section{Related Work}
\label{sec:related}

\subsection{Image Denoising Methods}
\label{ssec:img_den}


Denoising is a long-studied field with its roots in traditional signal processing. Since then, numerous methods have been proposed for image/video denoising \cite{img_den_survey0, img_den_survey1, img_den_survey2}. Prior to the advent of CNNs in vision tasks, image denoising methods either involved exploiting image priors \cite{classical1}, or using spatial filtering \cite{filter, filter2}.

With the introduction of CNNs in the vision domain, novel denoising methods demonstrated superior results compared to the previous classical methods \cite{img_den, img_den2, den_autoenc}. More recently, attention and transformer-based methods \cite{img_den_atten, restormer} have been proposed, achieving excellent performance.


\subsection{Video Denoising Methods}
\label{ssec:vid_den}

\begin{figure}[t!]
	\centering
	\includegraphics[width=\linewidth]{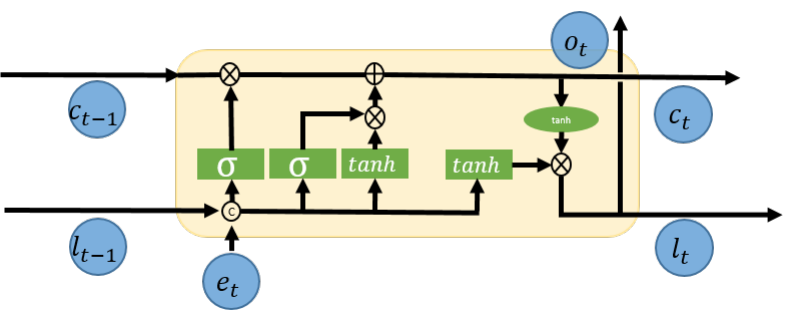}
	\caption{Sequential flow and gating mechanisms in an LSTM layer.}
	\label{fig:lstm}
\end{figure}

Unlike images, videos have a temporal dimension as well. The strong correlation that exists along this temporal axis in videos can be exploited for enhancing restoration quality. Therefore, work in this domain is focused on modeling spatio-temporal information rather than solely focusing on spatial features. 
Early work in video denoising explicitly modeled temporal information by utilising block matching \cite{vbm4d} and optical flow \cite{flow1, dvd, meshflow} to align frames before denoising. However, motion estimation and compensation methods are computationally demanding and are prone to accuracy issues.

To overcome the drawback of high computational complexity associated with spatial domain motion modeling, multi-frame based methods have been proposed \cite{udvd, rvidenet, fastdvd, derfs, videnn}, which take multiple consecutive video frames as input and output a single denoised frame. These methods attempt to fuse the information present in consecutive frames, thereby gaining some implicit temporal context for denoising. However, multi-frame based methods suffer from limitations in modeling temporal context since they lack explicit long term temporal modeling. Moreover, each frame needs to be processed multiple times, making these methods computationally inefficient. 

Recurrent Neural Networks (RNNs) gained the attention of the research community due to their ability in learning temporal context \cite{rnn_nlp1, rnn_nlp2}. A number of methods have made use of these recurrent networks for the denoising task \cite{emvd, deep_burst_den, recurr_res, latent}. While multi-frame methods only have a limited temporal memory, recurrent networks can effectively leverage long-term relationships between past frames and the current frame to enhance restoration quality.
More recently, encoder-decoder networks with attention mechanisms \cite{atten1} and transformers \cite{vrt, rvideformer} have been introduced, achieving excellent performance at the expense of high computational complexity. Transformers require a large amount of data for training and are computationally inefficient, leading to high training and inference times.

The majority of methods reported in contemporary literature suffer from very high complexities (with GFLOPs in the order of thousands) for processing a single frame. On the contrary, existing lightweight methods \cite{emvd, mobile} significantly lack denoising performance when compared to their heavier counterparts. Therefore, there is a need to develop lightweight models that can be run with limited computational resources while achieving better denosing quality. 

\subsection{Long Short-Term Memory}
\label{ssec:lstm}

Conventional Recurrent Neural Networks (RNNs) are constrained in their ability to model extensive temporal dependencies, thus limiting the incorporation of a substantial long-term temporal context. To address this limitation, Long Short-Term Memory (LSTM) \citep{lstm} networks were introduced. While initially only being used for Natural Language Processing (NLP) tasks such as machine translation \cite{translation} and language modeling \cite{modelling}, LSTMs have since found utility across a spectrum of vision-related endeavors \cite{captioning, sequencer}. 

Additionally, there have been endeavors to enhance the LSTM framework by integrating convolutional kernels, as demonstrated in prior work \cite{convlstm}. We depict the functioning of an LSTM layer in Figure~\ref{fig:lstm}. 

\begin{figure*}[t!]
	\centering
	\includegraphics[width=0.9\linewidth]{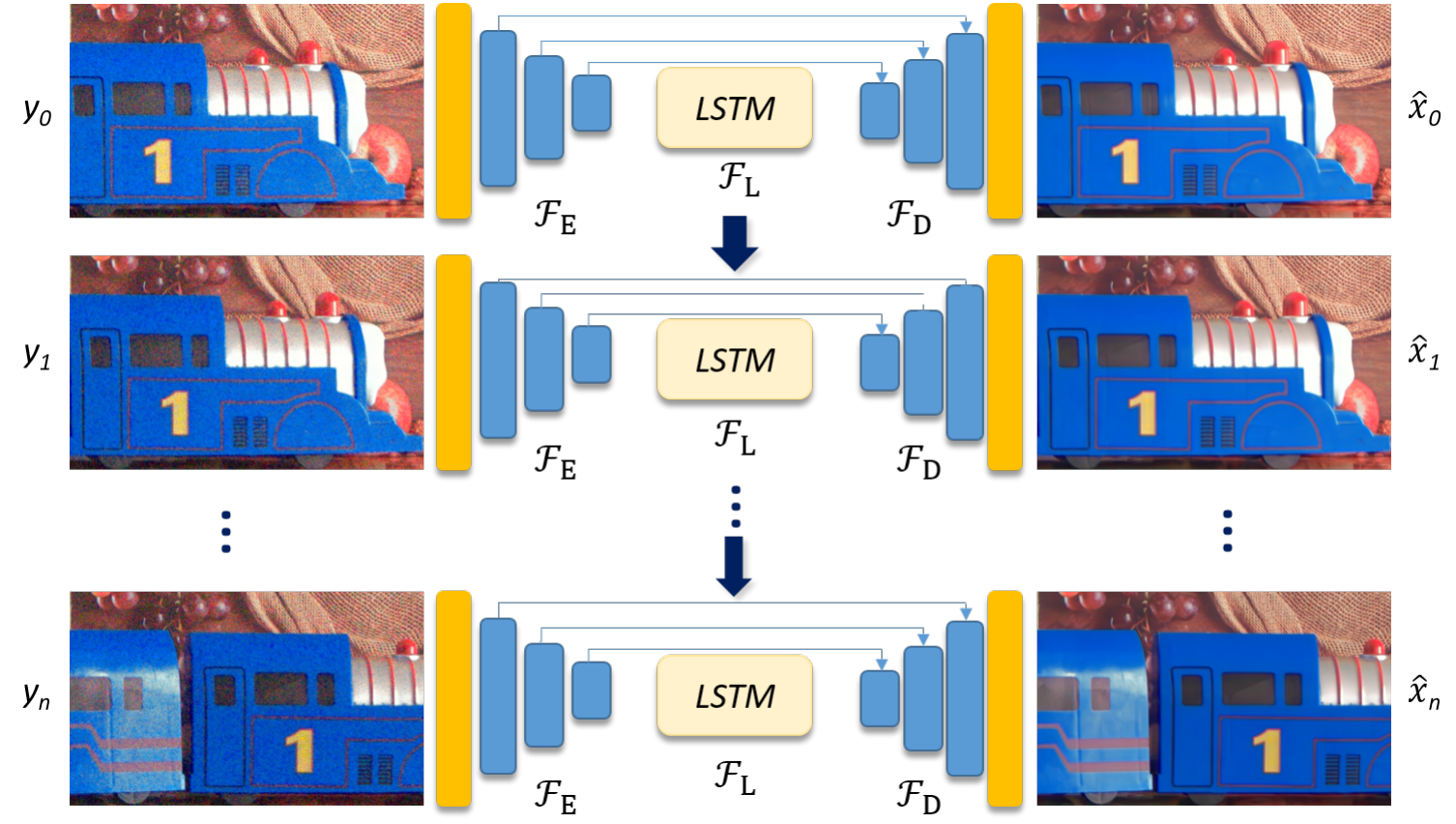}
	\caption{An overview of our proposed method.} 
	\label{fig:init}
\end{figure*}

\section{Problem Formulation}
\label{sec:formulation}

Consider a degraded video sequence, denoted as $\bm{Y}_{T} = \{y_1, y_2, ..., y_T\}$, comprising $T$ frames. Here, $y_t \in \mathbb{R}^{H \times W \times C}$ is the frame at time $t$ characterized by its height $H$, width $W$ and channels $C$. Similarly, we define the corresponding set of clean frames as  $\bm{X}_T = \{x_1,x_2, ..., x_T\}$ which undergo degradation to yield the observable frames $\bm{Y}_T$, such that, 
\begin{equation}
	y_t = f(d(x_t) + n_t).
\end{equation}
In this context, the function $d$ encapsulates degradations introduced during capture (eg. motion blur, over-exposure, etc), $n_t$ accounts for noise due to the camera sensor, and $f$ encompasses transformations that transpire post-capture (eg. ISP, compression, etc.). Notably, our focus in this work is on a specific scenario where $d(.)$ is an identity operation, leading to the introduction of degradation solely through noise. Moreover, we investigate two distinct scenarios for the function $f(.)$: (i) RAW domain restoration, characterized by $f(.)$ as identity, yielding $y_t = x_t + n_t$, and (ii) RGB domain restoration, wherein $f(.)$ represents the ISP pipeline of the capturing device i.e.  $y_t = f(x_t + n_t)$. Our objective is to construct the function $\bm{\mathcal{F}}: \mathbb{R}^{T \times H \times W \times C} \to \mathbb{R}^{T \times H \times W \times C}$, which can effectively recover the clean frames denoted as $\bm{X_T}$, such that $\bm{\hat{X}}_T = \mathcal{F}(\bm{Y}_T) \approx f(\bm{X_T}) $ for RGB denoising, and $\bm{\hat{X}}_T = \mathcal{F}(\bm{Y}_T) \approx \bm{X_T} $ for RAW denoising.  

\section{Proposed Approach}
\label{sec:approach}
The restoration function $\bm{\mathcal{F}}$ detailed in Section \ref{sec:formulation} is constructed through the composition of three distinct functions: a Spatial Encoder ($\bm{\mathcal{F}_E}$), a Recurrence Block ($\bm{\mathcal{F}_L}$), and a Spatial Decoder($\bm{\mathcal{F}_D}$),
\begin{equation}
	\label{eq:comp}
	\bm{\mathcal{F}} = \bm{\mathcal{F}_E} \circ \bm{\mathcal{F}_L} \circ \bm{\mathcal{F}_D}.
\end{equation}

\subsection{Encoded LSTMs}

The Spatial Encoder and Decoder are instrumental in extracting and reconstructing the fine spatial details of video frames. In tandem, an LSTM model, embedded within the encoded latent space, adeptly captures the temporal dynamics. The incorporation of LSTM is crucial in understanding both the immediate and extended temporal relationships between frames, ensuring the production of smooth, flicker-free videos. While past approaches have explored a dual branch approach, segregating the extraction of spatial and temporal features \cite{dualbranch}, our innovation resides in an integrated spatio-temporal approach. Significantly efficient and lightweight, our proposed model is optimally suited for on-device video denoising.

We employ the compositional framework outlined in equation \ref{eq:comp} to realize the denoising function $\bm{\mathcal{F}}$ for enhancing captured videos. Given a noisy video sequence $\bm{Y}_T$, the initial step involves the encoder extracting spatially encoded features for each frame, 
\begin{equation}
	\label{eq:encoder}
	e_t = \bm{\mathcal{F}_E} (y_t) , \quad \forall t \in \{1, ..., T\}.
\end{equation} 

Subsequently, a recursive relationship is established among the spatially encoded frames $E_T = \{e_1, ..., e_T\}$, facilitated by an LSTM model. This LSTM model focuses on capturing the temporal dynamics within $E_T$, leading to the creation of the `spatio-temporal encodings' $L_T = \{l_1,...,l_T\}$, 
\begin{equation}
	\label{eq:LSTM}
	l_t = \bm{\mathcal{F}_L}(e_t, l_{t-1}, c_{t-1}), \quad \forall t \in \{1, ..., T\},
\end{equation} 
where, $e_t$ denotes the spatially encoded frame at time $t$, $l_{t-1}$ denotes the LSTM output from the preceding time step $(t-1)$, and $c_{t-1}$ signifies the LSTM cell state at time $(t-1)$. Initialization at $t=0$ involves setting $l_{t-1}$ and $c_{t-1}$ to 0. 

Finally, the `spatio-temporal encodings' $L_t$ undergo decoding through $\bm{\mathcal{F}_D}$ to reconstruct the pristine video frame as,
\begin{equation}
	\label{eq:decoder}
	\hat{x}_t = \sigma(\bm{\mathcal{F}_D}(l_t)), \quad \forall t \in \{1,...,T\},
\end{equation}
where $\sigma$ denotes the sigmoid activation function. A broad overview of our proposed approach can be found in Figure ~\ref{fig:init}.

\subsection{Model Architecture}
\label{ssec:architecture}
\begin{figure*}[ht!]
	\centering
	\includegraphics[width=0.9\linewidth]{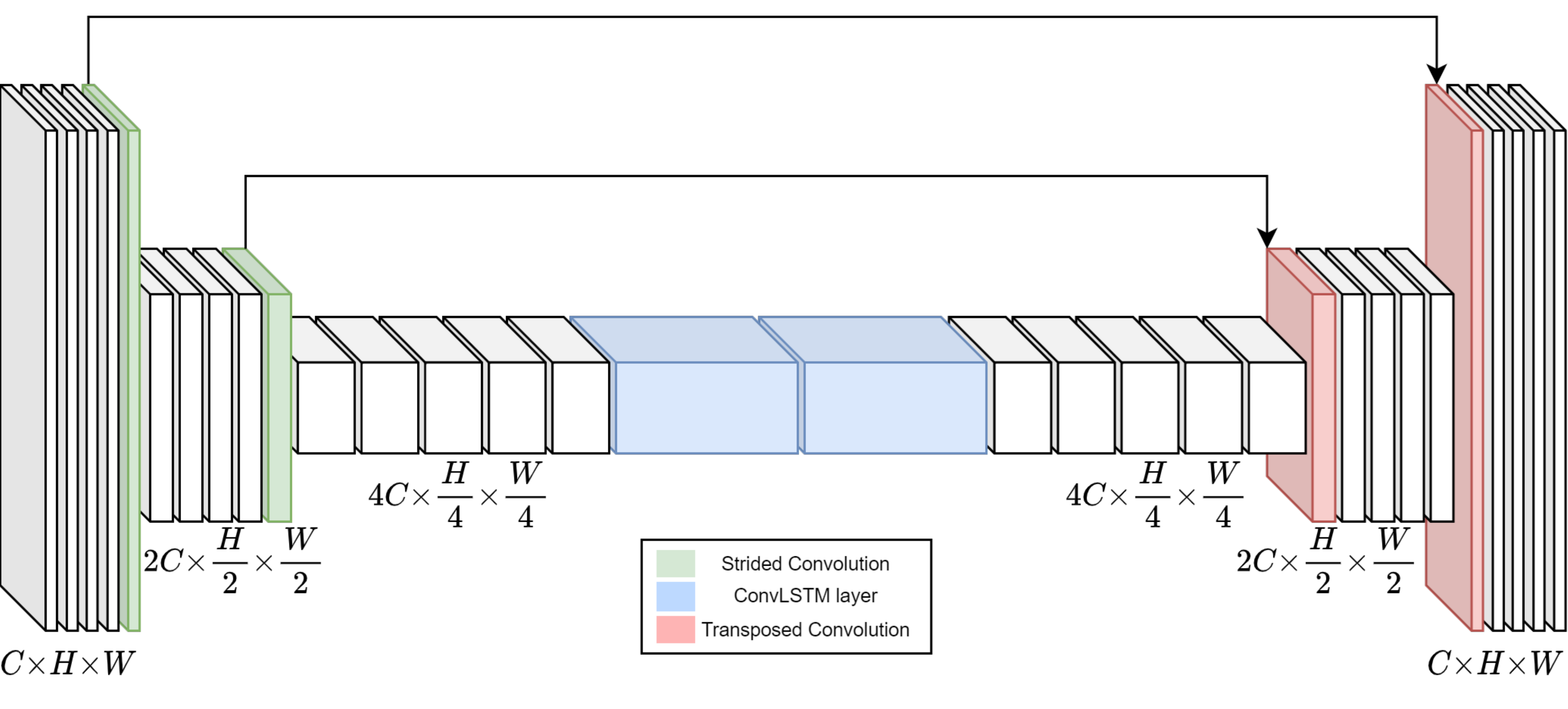}
	\caption{Detailed overview of our model architecture for a single frame. We follow a U-Net style symmetric Encoder-Decoder architecture with skip connections. The encoder and decoder can be broken down into three stages each, with each stage consisting of 5 convolutional layers. The last (first) layer of each stage in the encoder (decoder) performs downsampling (upsampling) using strided convolutions (transposed convolutions). The input video frame and denoised output frame are also connected by a residual connection (not shown).} 
	\label{fig:model}
\end{figure*}

We employ a U-Net inspired architecture for approximating the functions of the Encoder and the Decoder. The input frame at time $t$ undergoes initial processing through the Encoder function, $\bm{\mathcal{F}_E}: \mathbb{R}^{W \times H \times C} \to \mathbb{R}^{w \times h \times c}$, structured with three Encoder blocks, each made up of five convolutional layers. Consecutive blocks employ strided convolutions to halve the resolution, resulting in the latent space configuration of $h = H/4$ and $w = W/4$. Subsequently, the recurrent block in the latent space, $\bm{\mathcal{F}_L}: \mathbb{R}^{w \times h \times c} \to \mathbb{R}^{w \times h \times c}$, is implemented through a 2-layer LSTM. The choice of Convolutional-LSTM \cite{convlstm} architecture is favored over its Fully Connected counterpart due to its convolutional nature, enabling the acquisition of both spatial and temporal information. Finally, the Decoder function, $\bm{\mathcal{F}_D}: \mathbb{R}^{w \times h \times c} \to  \mathbb{R}^{W \times H \times C}$, is responsible for transforming the spatio-temporally encoded features back into the image space. The Decoder is symmetric in design to the Encoder, with each Encoder block connected to its corresponding Decoder block via a residual connection. The entire architectural arrangement is succinctly depicted in Figure \ref{fig:model}. 

To further enhance model efficiency, we introduce a Pixel Unshuffle/Shuffle layer before/after the model's first/last layer, successfully reducing the model complexity by 75\%. We refer to the model employing Pixel Shuffle as LLVD-S (Small) and the version without it as LLVD-L (Large). In Section \ref{sec:exp}, we comprehensively evaluate these models. 

\subsection{Loss Function Design}
\label{ssec:losses}
The proposed model is guided towards learning the video denoising function, $\bm{\mathcal{F}}$, through the minimization of a composite loss function combining $L_1$ and $L_2$ distances between the generated denoised video and the Ground Truth. The concurrent use of $L_1$ and $L_2$ loss serves to capture both the error magnitude and direction in the model predictions, thereby striking a balance between precision and robustness. While $L_2$ emphasizes precision and heavily penalizes large errors, it can be very sensitive to outliers. The incorporation of $L_1$ loss fosters outlier robustness and generates sparse solutions, yielding visually appealing sharper denoised videos. Furthermore, we enhance structural integrity by maximizing the Structural Similarity (SSIM) metric between denoised videos and the Ground Truth. Consequently, the loss function to be minimized is formulated as follows, 
\begin{gather}
	\begin{align}
		\nonumber \min_{\bm{\mathcal{F}}(.)} \mathcal{L}(\bm{Y}_T) = & ||\bm{\mathcal{F}}(\bm{Y}_T) - \bm{X}_T||_2^2 + \lambda_1 ||\bm{\mathcal{F}}(\bm{Y}_T) - \bm{X}_T||_1 \\
		&+ \lambda_2 [1 - SSIM(\bm{\mathcal{F}}(\bm{Y}_T), \bm{X}_{T})],
	\end{align}
\end{gather}
where $\lambda_1$ and $\lambda_2$ are hyper-parameters governing the influence of individual loss terms. In our experimental setup, we set $\lambda_1=0.1$ and $\lambda_2=0.01$ to appropriately control these contributions.

\section{Experiments}
\label{sec:exp}


We present two versions of our proposed architecture, LLVD-S (Small) and LLVD-L (Large). To evaluate our method, we use RGB images with synthetic noise and RGB+RAW images with captured noise. Our model's performance is evaluated against several SOTA video denoising techniques, utilizing Peak Signal-to-Noise Ratio (PSNR) and Structural Similarity Index Measure (SSIM) as evaluation metrics. Finally, we also provide a comparative analysis of the computational complexities associated with these methods, quantified in terms of Giga FLoating point OPerations (GFLOPs).

The assessment is conducted on scenes sourced from the Captured Raw Video Dataset (CRVD) \cite{rvidenet} for real-world sensor noise introduced during capture, DAVIS \cite{davis} and Set8 \cite{dvd} for synthetic additive noise.


\begin{figure*}[ht!]
	\centering 
	
	
 
 	\subfloat[Set8, $\sigma = 30$]{%
		\includegraphics[clip,width=0.95\linewidth]{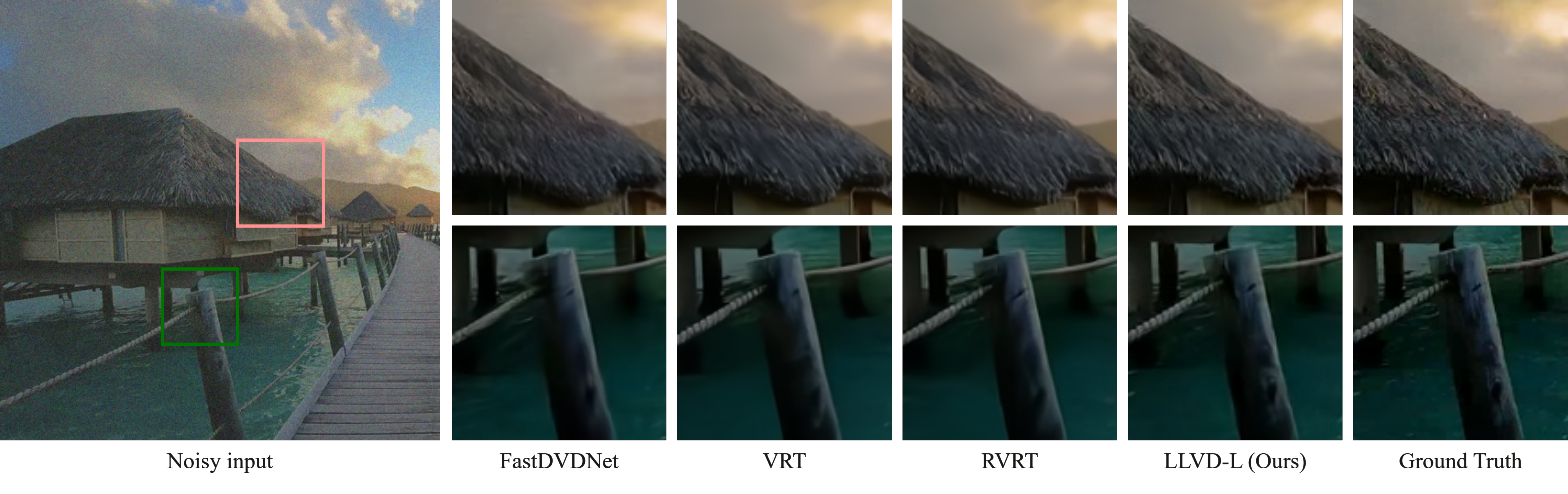}%
		\label{fig:quali1}
	}

	\subfloat[CRVD, ISO 25600]{%
		\includegraphics[clip,width=0.95\linewidth]{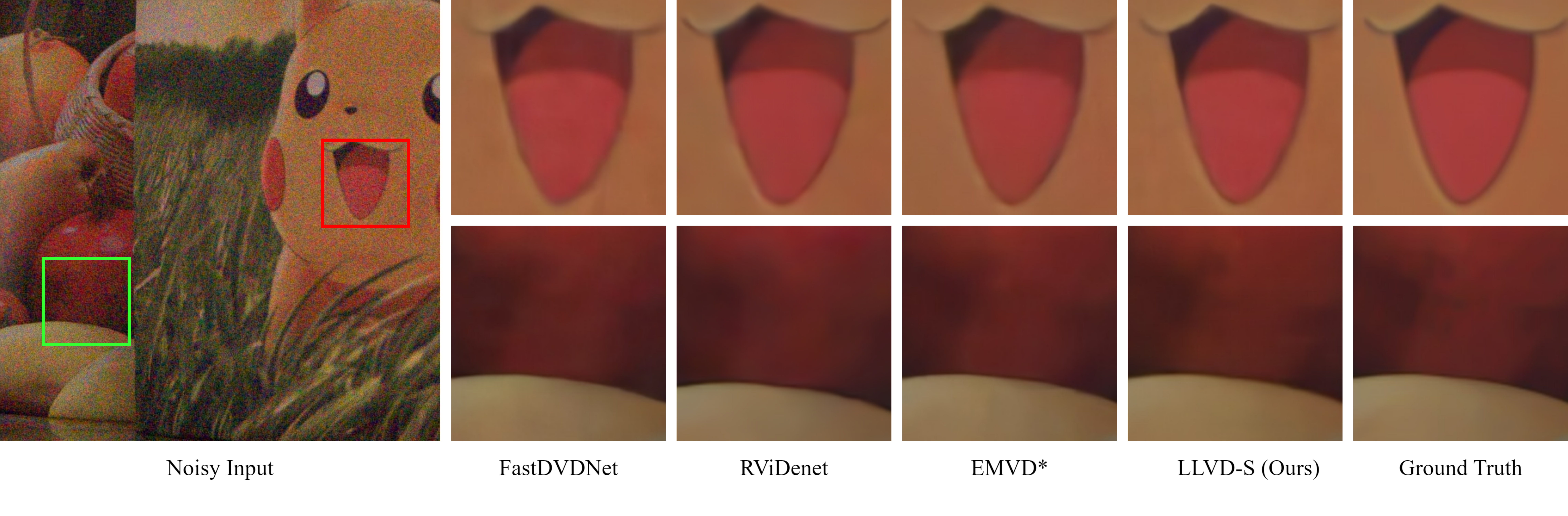}%
		\label{fig:quali2}
	}	
	
	\caption{Qualitative comparison on scenes in the CRVD and Set8 testsets. Zoom in for better observation.}
	\label{fig:quali}
\end{figure*}

\subsection{Training Details}
\label{ssec:details}

\textbf{Real-World Noise:} In accordance with \cite{rvidenet}, the initial phase of training uses the Supervised Raw Video Denoising (SRVD) dataset. We employ the Adam optimizer, setting the learning rate at $10^{-4}$ and batch size to $4$. Subsequently, the model is finetuned on scenes 1 through 6 from the CRVD dataset. Evaluation is conducted on scenes 7 to 11 of the same dataset.

\textbf{Synthetic Noise:} Mirroring the data preparation methodology from FastDVDNet \cite{dvd, fastdvd}, we generate synthetically degraded videos by introducing Additive White Gaussian Noise (AWGN) to videos in the DAVIS training set. Our model's performance is evaluated on the DAVIS and Set8 testsets, with noise levels of $\sigma \in [10,20,30,40,50]$.

For both noise models, our network processes $25$ frames from each video segment. For the DAVIS dataset, we train on full-resolution (480p) videos, whereas for the CRVD dataset we use randomly cropped $128\times 128$ patches. When processing raw videos, we maintain the CFA Bayer Pattern during spatial cropping as recommended in \cite{bayer_preserve}. To address the limited 7 frame scope of the CRVD dataset, we extend the training sequence to 25 frames through mirroring. Such an augmentation technique enhances the temporal context for improved learning. Our model is trained using two NVIDIA A100 GPUs. 

%

\subsection{Evaluation on Real-world Noise}
\label{ssec:crvd_results}
We conduct a comparative analysis of our method against various video denoising techniques, including  FastDVDNet \cite{fastdvd}, RViDenet \cite{rvidenet}, EMVD \footnote{Since the code for EMVD has not been released, we use an unofficial implementation: \href{https://github.com/Baymax-chen/EMVD/}{https://github.com/Baymax-chen/EMVD/}.} \cite{emvd}, BSVD \cite{bsvd}, and RViDeformer \cite{rvideformer}. A comprehensive quantitative comparison is showcased in Table~\ref{table:main_results}. We also provide a concise comparison of LLVD-S/L against other methods in Figure~\ref{fig:graph}.

Our experiments reveal that our lighter model, LLVD-S, demonstrates significant superiority over its heavier counterparts in the sRGB domain. Specifically, on the sRGB CRVD testset, LLVD-S achieves performance parity with RViDeformer-T \cite{rvideformer} ($\sim2.5\times$ heavier than LLVD-S) and surpasses BSVD-24 with a marginal PSNR improvement of 0.12dB, while concurrently reducing computations by a factor of $6$ (See Table~\ref{table:main_results}). 

For our heavier model, LLVD-L (obtained by removing the pixel shuffle operations), we observe superior denoising prowess over RViDeformer-M ($\sim2.5\times$ heavier than LLVD-L) by 0.3dB on the raw CRVD testset and 0.1dB on the sRGB CRVD testset. 
The improved performance of LLVD-S/L over RViDeformer can be attributed to the effectiveness of LSTM in modeling temporal context in latent space while being computationally efficient.

\subsection{Evaluation on Synthetic Noise}
\label{ssec:davis_results}

We also conduct an extensive evaluation of LLVD's effectiveness on the DAVIS and Set8 datasets, utilizing synthetic AWGN noise. We benchmark LLVD against a variety of blind and non-blind denoising techniques, namely FastDVDNet \cite{fastdvd}, BP-EVD \cite{bpevd}, BSVD \cite{bsvd}, UDVD \cite{udvd}, FloRNN \cite{flornn}, PaCNet \cite{pacnet}, PTFN \cite{ptfn}, VRT \cite{vrt}, RVRT \cite{rvrt}, RDRF \cite{rdrf}, and ReVid \cite{revid}. The comparative analysis is detailed in Table \ref{table:davis}.

Our heavier variant, LLVD-L, demonstrates remarkable performance on synthetic noise as well, surpassing all models of similar complexity. In comparison with the current SOTA methods such as VRT, RVRT, RDRF, and ReVid, LLVD-L exhibits a marginal decrease in performance (less than 1 dB) while requiring $10\times$ times fewer computations. Moreover, VRT, RVRT and ReViD are non-blind denoising techniques, which require the noise map as an input. LLVD is noise-blind and does not require any prior information about noise characteristics.

Additionally, we evaluate performance in a cross-dataset setting: evaluating our method on the Set8 testset when trained on the DAVIS dataset. In this setting, LLVD surpasses nearly all denoising techniques except ReViD, indicating the superior generalizability of our method. This cross-dataset performance reinforces LLVD's capacity for real-world, in-the-wild denoising scenarios.

In addition to our quantitative analysis, we provide visual results in Figure~\ref{fig:quali}. LLVD's ability to retain high frequency details better than other models can be observed in Figure~\ref{fig:quali1} and in the background areas of Figure~\ref{fig:quali2}. 


\renewcommand{\arraystretch}{1.15}
\definecolor{lyellow}{rgb}{1,1,0.5}

\begin{table}
	\centering	
	\begin{adjustbox}{width = {\columnwidth}}
	\begin{tabular}{cccc} 
		\thickhline
		\multirow{2}{*}{MODEL} & \multirow{2}{*}{GFLOPs} & sRGB~                          & RAW                            \\ 
		\cline{3-4}
		&                         & \multicolumn{1}{l}{PSNR/SSIM} & \multicolumn{1}{l}{PSNR/SSIM}  \\ 
		\hline
		FastDVDNet-S           & 22.16                   & 37.43/0.969                  	 & 42.25/0.980                    \\
		EMVD-S                 & 5.38                    & 38.27/0.972                  	 & 42.63/0.985                    \\
		FastDVDNet             & 664.99                  & 39.91/0.981                  	 & 44.30/0.989                    \\
		EMVD-M                 & 79.52                   & 39.53/0.979                  	 & 44.05/0.989                    \\
		RViDeNet               & 2079.74                 & 39.95/0.979                  	 & 43.97/0.987                    \\
		BSVD-16                & 78.76                   & 40.17/0.980                   	 & 44.10/0.988                    \\
		BSVD-24                & 175.46                  & \underline{40.48}/\textbf{0.982}	 & \textbf{44.39}/\underline{0.989}\\
		RViDeformer-T          & 77.62                   & \textbf{40.60}/\underline{0.981}	 & \underline{44.34}/0.988		  \\
		\rowcolor{lyellow} LLVD-S                 & \textbf{29.96}                   & \textbf{40.60}/\textbf{0.982}	 & 43.61/\textbf{0.993}  		  \\ 
		\hline
		EMVD-L                 & 2542.86                 & --				            	 & 44.51/0.989                    \\
            BP-EVD          & 131                   & --              	     & 44.42/0.988                    \\
		RViDeformer-S          & 142.6                   & 41.00/\underline{0.983}              	     & 44.65/0.989                    \\
		RViDeformer-M          & 287.34                  & \underline{41.29}/\textbf{0.984}	 	         & \underline{44.89}/\underline{0.990}                    \\
		\rowcolor{lyellow} LLVD-L                 & \textbf{117.4}                   & \textbf{41.41/\textbf{0.984}}     & \textbf{45.18/\textbf{0.996}}  \\
		\thickhline
	\end{tabular}
	\end{adjustbox}

 	\caption{Comparison of our method (LLVD-S/L) against various SOTA video denoising methods. Best results are shown in \textbf{bold}, second best are \underline{underlined}. The results of FastDVDNet-S and FastDVDNet are quoted from \cite{emvd}.}
	\label{table:main_results}
\end{table}

\begin{table*}
\centering
\label{table:davis}
\resizebox{\linewidth}{!}{
\begin{tabular}{ccccccccccccccc}

\thickhline
\multicolumn{1}{l}{\multirow{2}{*}{}} & \multirow{2}{*}{MODEL} & \multirow{2}{*}{GFLOPS} & \multicolumn{6}{c}{DAVIS} & \multicolumn{6}{c}{SET8} \\ 
\cline{4-15}
\multicolumn{1}{l}{} &  &  & 10 & 20 & 30 & 40 & 50 & AVG & 10 & 20 & 30 & 40 & 50 & AVG \\ 
\hline
\multirow{9}{*}{\rotatebox{90}{NON-BLIND}} & FastDVDNet & 263.20 & 38.71 & 35.77 & 34.04 & 32.82 & 31.86 & 34.64 & 36.44 & 33.43 & 31.68 & 30.46 & 29.53 & 32.31 \\
 & BP-EVD & 131 & 39.13 & 36.04 & 34.24 & 32.99 & 31.97 & 34.87 & 37.33 & 33.95 & 32.04 & 30.73 & 29.73 & 32.76 \\
 & FloRNN & 375.25 & 40.16 & 37.52 & 35.89 & 34.66 & 33.67 & 36.38 & 37.57 & 34.67 & 32.97 & 31.75 & 30.8 & 33.55 \\
 & PaCNet & 117.42 & 39.97 & 37.1 & 35.07 & 33.57 & 32.39 & 35.62 & 37.06 & 33.94 & 32.05 & 30.7 & 29.66 & 32.68 \\
 & BSVD-64 & 486.35 & 39.81 & 36.82 & 35.09 & 33.86 & 32.91 & 35.70 & 36.74 & 33.83 & 32.14 & 30.97 & 30.06 & 32.75 \\
 & PTFN & 111.45 & 39.86 & 37.05 & 35.41 & 34.24 & 33.24 & 35.96 & 36.82 & 33.99 & 32.36 & 31.22 & 30.34 & 32.95 \\
 & VRT & 6287.44 & \uline{40.82} & \uline{38.15} & 36.52 & 35.32 & 34.36 & 37.03 & \uline{37.88} & \uline{35.02} & \uline{33.35} & 32.15 & 31.22 & \uline{33.92} \\
 & RVRT & 515.44 & 40.57 & 38.05 & \uline{36.57} & \uline{35.47} & \uline{34.57} & \uline{37.05} & 37.53 & 34.83 & 33.3 & \uline{32.21} & \uline{31.33} & 33.84 \\
 & ReViD & 182.8 & \textbf{41.11} & \textbf{38.61} & \textbf{37.1} & \textbf{35.98} & \textbf{35.08} & \textbf{37.58} & \textbf{38.07} & \textbf{35.41} & \textbf{33.87} & \textbf{32.76} & \textbf{31.88} & \textbf{34.4} \\ 
\hline
\multirow{5}{*}{\rotatebox{90}{BLIND}} & UDVD & OOM & – & – & 33.86 & 32.61 & 31.63 & 32.7 & – & – & 32.01 & 30.82 & 29.89 & 30.91 \\
 & RDRF & – & 39.54 & 36.4 & 34.55 & 33.23 & 32.2 & 35.18 & \uline{36.67} & \uline{34.00} & \uline{32.39} & \uline{31.23} & \uline{30.31} & \uline{32.92} \\
 & BSVD-64 & 486.35 & 39.68 & 36.66 & 34.91 & 33.68 & 32.72 & 35.53 & 36.54 & 33.70 & 32.02 & 30.85 & 29.95 & 32.61 \\
 & PTFN & \textbf{111.45} & \uline{39.79} & \uline{36.96} & \uline{35.31} & \uline{34.14} & \uline{33.22} & \uline{35.88} & 36.64 & 33.89 & 32.28 & 31.14 & 30.26 & 32.84 \\
 & LLVD-L & \uline{116.5} & \textbf{41.81} & \textbf{37.93} & \textbf{35.8} & \textbf{34.36} & \textbf{33.29} & \textbf{36.64} & \textbf{37.5} & \textbf{35.13} & \textbf{33.55} & \textbf{32.38} & \textbf{31.48} & \textbf{34.01} \\
\thickhline
\end{tabular}
}
\caption{Comparison of our method (LLVD-L) against various SOTA video denoising methods on the DAVIS and Set8 testsets. Best results are shown in \textbf{bold}, second best are \underline{underlined}. The GFLOP values have been calculated for denoising a single frame of $854\times 480$ resolution. The complexity value for ReViD is for a single $256\times 256$ frame as quoted from their paper. At the same resolution, LLVD's compelxity is $18.76$ GFLOPs, $10\times$ lighter. The code for ReViD and RDRF is not publicly available.}

\label{table:davis}
\end{table*}

        

\subsection{Ablation Study}
\label{ssec:ablation}

As elaborated in Section~\ref{sec:approach}, our model is composed of three functions ($\mathcal{F}_E, \mathcal{F}_L, \text{ and } \mathcal{F}_D$). In our ablation study, we evaluate the impact of these functions on the denoising performance. The training methodology remains consistent with Section~\ref{ssec:details}. Table~\ref{table:ablation} depicts the outcomes of our ablation study.

First, we eliminate all temporal learning elements from the model, rendering it purely spatial. The ConvLSTM layers $\mcal{F}_L$ are removed, leaving only the encoder-decoder network ($\mcal{F}_E \text{ and } \mcal{F}_D$). This configuration essentially translates to an image denoising network applied individually to each frame of the video. As anticipated, we observe a substantial drop in the PSNR values, showcasing the indispensability of temporal learning in video data processing. Subsequently, we introduce a single ConvLSTM layer between the encoder-decoder components. Notably there is an improvement over the purely spatial network. Nevertheless, the capacity of a solitary ConvLSTM layer proves inadequate for effective temporal feature extraction. The proposed model with 2 LSTM layers further improves the performance and is optimal for temporal feature extraction. 

By completely excluding the encoder-decoder the network is comprised solely of two ConvLSTM layers. We keep the number of channels constant through both layers. Although this arrangement retains spatial learning (due to the use of ConvLSTM layers instead of FC-LSTM layers), the capacity for such learning is markedly constrained, explaining the low PSNR value.


\begin{table}
	\centering
	\caption{Ablation study, computed on the sRGB CRVD testset.}
	\begin{adjustbox}{width = {\columnwidth}}
	\begin{tabular}{cccccc} 
		\thickhline
		Pixel Shuffle             & $\times$   & \checkmark & \checkmark & \checkmark & $\times$    \\
		Enc-Dec                   & $\times$   & \checkmark & \checkmark & \checkmark & \checkmark  \\
		LSTM (\textbackslash{}\#) & 2          & 0          & 1         & 2           & 2          \\ 
		\hline
		GFLOPs                    & 1.50       & 19.58       & 24.76     & 29.96       & 117.40     \\ 
		\hline
		PSNR                      & 35.58      & 38.26      & 39.96     & 40.60       & 41.41      \\
		SSIM                      & 0.894      & 0.968      & 0.977     & 0.982       & 0.984      \\
		\thickhline
	\end{tabular}
	\end{adjustbox}
	\label{table:ablation}
\end{table}

\section{Conclusion}
\label{sec:conclusion}

In this paper, we proposed a novel approach for Video Restoration, with a focus on denoising in both RAW and RGB domains. The Latent LSTM Video Denoiser (LLVD) demonstrated remarkable performance performance improvements in denoising video frames with both captured and synthetic noise. The implementation of LSTM blocks in the encoded latent space serves as a robust framework for capturing both short-term and long-term temporal relationships without requiring a high computational complexity. Through extensive experimentation, we showcased the superiority of LLVD-L and LLVD-S (a lighter variant of our model for deployment in mobile devices), outperforming heavier models while maintaining reduced computational burden. Our ablation study further underscores the critical importance of each component in our model, emphasizing the significance of temporal modeling for video restoration. Our comprehensive comparison against SOTA denoising methods, as well as the evaluation of both LLVD-L and LLVD-S variants, solidifies the credibility and practicality of the proposed approach. In conclusion, the Latent LSTM Video Denoiser (LLVD) emerges as a promising solution for video denoising on resource constrained devices such as mobiles, boasting remarkable denoising capabilities, efficiency gains, and a versatile framework for capturing intricate temporal relationships.


\bibliography{root}

\bibliographystyle{icml2024}

%
%

\end{document}